\documentclass[letterpaper, 10 pt, conference]{ieeeconf}
\IEEEoverridecommandlockouts                              
\usepackage{comment}
\usepackage{url}

\usepackage{multirow}
\usepackage{graphicx}
\usepackage{epsfig}
\usepackage{epic,eepic}
\usepackage{times}
\usepackage{mathptmx}
\usepackage{cite}
\usepackage{color}

\usepackage{times}

\definecolor{red}{rgb}{1,0,0}
\definecolor{green}{rgb}{0,1,0}
\definecolor{blue}{rgb}{0,0,1}
\definecolor{violet}{rgb}{1,0,1}
\definecolor{cyan}{cmyk}{1,0,0,0}
\definecolor{magenta}{cmyk}{0,1,0,0}
\definecolor{yellow}{cmyk}{0,0,1,0}

\definecolor{white}{rgb}{1,1,1}

\usepackage{arydshln}

\newcommand{\CO}[1]{}

\newcommand{\CommentOut}[1]{}

 \newcommand{\editage}[1]{}

\newcommand{\islarge}{}

\begin{document}


\newcommand{\FIG}[3]{
\begin{minipage}[b]{#1cm}
\begin{center}
\includegraphics[width=#1cm]{#2}\\
{\scriptsize #3}
\end{center}
\end{minipage}
}

\newcommand{\FIGU}[3]{
\begin{minipage}[b]{#1cm}
\begin{center}
\includegraphics[width=#1cm,angle=180]{#2}\\
{\scriptsize #3}
\end{center}
\end{minipage}
}

\newcommand{\FIGm}[3]{
\begin{minipage}[b]{#1cm}
\begin{center}
\includegraphics[width=#1cm]{#2}\\
{\scriptsize #3}
\end{center}
\end{minipage}
}

\newcommand{\FIGR}[3]{
\begin{minipage}[b]{#1cm}
\begin{center}
\includegraphics[angle=-90,width=#1cm]{#2}
\\
{\scriptsize #3}
\vspace*{1mm}
\end{center}
\end{minipage}
}

\newcommand{\FIGRpng}[5]{
\begin{minipage}[b]{#1cm}
\begin{center}
\includegraphics[bb=0 0 #4 #5, angle=-90,clip,width=#1cm]{#2}\vspace*{1mm}
\\
{\scriptsize #3}
\vspace*{1mm}
\end{center}
\end{minipage}
}

\newcommand{\FIGCpng}[5]{
\begin{minipage}[b]{#1cm}
\begin{center}
\includegraphics[bb=0 0 #4 #5, angle=90,clip,width=#1cm]{#2}\vspace*{1mm}
\\
{\scriptsize #3}
\vspace*{1mm}
\end{center}
\end{minipage}
}

\newcommand{\FIGpng}[5]{
\begin{minipage}[b]{#1cm}
\begin{center}
\includegraphics[bb=0 0 #4 #5, clip, width=#1cm]{#2}\vspace*{-1mm}\\
{\scriptsize #3}
\vspace*{1mm}
\end{center}
\end{minipage}
}

\newcommand{\FIGtpng}[5]{
\begin{minipage}[t]{#1cm}
\begin{center}
\includegraphics[bb=0 0 #4 #5, clip,width=#1cm]{#2}\vspace*{1mm}
\\
{\scriptsize #3}
\vspace*{1mm}
\end{center}
\end{minipage}
}

\newcommand{\FIGRt}[3]{
\begin{minipage}[t]{#1cm}
\begin{center}
\includegraphics[angle=-90,clip,width=#1cm]{#2}\vspace*{1mm}
\\
{\scriptsize #3}
\vspace*{1mm}
\end{center}
\end{minipage}
}

\newcommand{\FIGRm}[3]{
\begin{minipage}[b]{#1cm}
\begin{center}
\includegraphics[angle=-90,clip,width=#1cm]{#2}\vspace*{0mm}
\\
{\scriptsize #3}
\vspace*{1mm}
\end{center}
\end{minipage}
}

\newcommand{\FIGC}[5]{
\begin{minipage}[b]{#1cm}
\begin{center}
\includegraphics[width=#2cm,height=#3cm]{#4}~$\Longrightarrow$\vspace*{0mm}
\\
{\scriptsize #5}
\vspace*{8mm}
\end{center}
\end{minipage}
}

\newcommand{\FIGf}[3]{
\begin{minipage}[b]{#1cm}
\begin{center}
\fbox{\includegraphics[width=#1cm]{#2}}\vspace*{0.5mm}\\
{\scriptsize #3}
\end{center}
\end{minipage}
}









\islarge{
\LARGE
}

\newcommand{\titleauthor}[2]{\title{\bf\Large%
#1}%
\author{#2}%
\maketitle%
}

\titleauthor{
CON:
Continual Object Navigation
via Data-Free Inter-Agent Knowledge Transfer
in Unseen and Unfamiliar Places
}{Kouki Terashima ~~~~~  Daiki Iwata ~~~~~ Kanji Tanaka ~~~~\thanks{Our work has been supported in part by JSPS KAKENHI Grant-in-Aid for Scientific Research (C) 20K12008 and 23K11270.}\thanks{$*$%
K. Terashima, D. Iwata, K. Tanaka are with Robotics Coarse, 
Department of Engineering, University of Fukui, Japan. 
{\tt\small{\{mf240271@g., mf240050@g., tnkknj@\}u-fukui.ac.jp}}}}

\maketitle

\newcommand{\tb}[1]{\textbf{#1}}

\newcommand{\figA}[1]{
\begin{figure}[t]
  \begin{center}
\FIG{8}{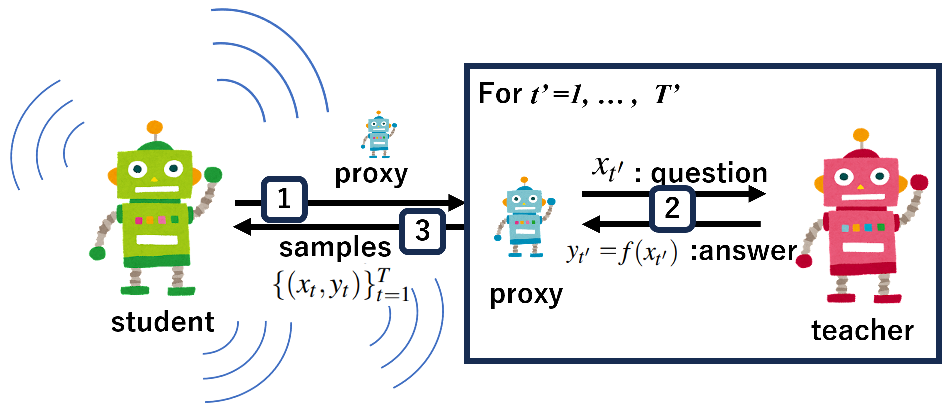}{}\\
\caption{CON formulation: 
We consider a continual learning (CL) problem where knowledge for object navigation (ON) tasks (e.g., object feature maps, neural state-action maps) is transferred from existing black-box models (teachers) to a new model (student). Unlike typical CL setups, a teacher-side plug-and-play knowledge transfer module called ``student proxy" (2) is allowed to engage in high-frequency question-and-answer sessions with the teacher, 
while the questions sent (1) and the responses received (3) are required to be lightweight and beneficial messages (e.g., language prompts).
The figure is adapted from \cite{tsukahara2024trainingselflocalizationmodelsunseen}.
}\label{fig:A}
  \end{center}
\end{figure}
}

\newcommand{\figB}{
\begin{figure}[t]
\begin{center}
\FIGR{8}{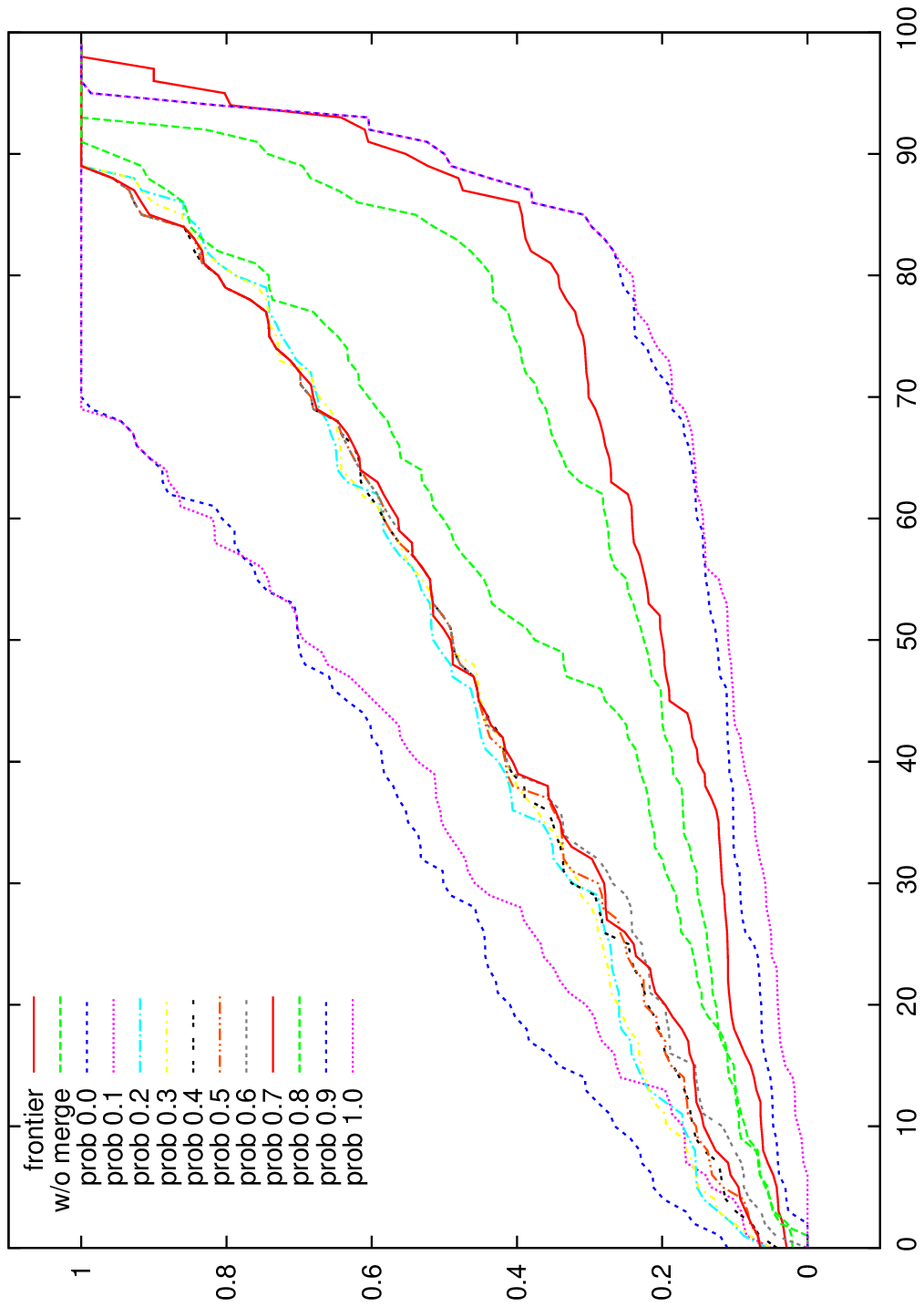}{Constrained start scenario}\\
\FIGR{8}{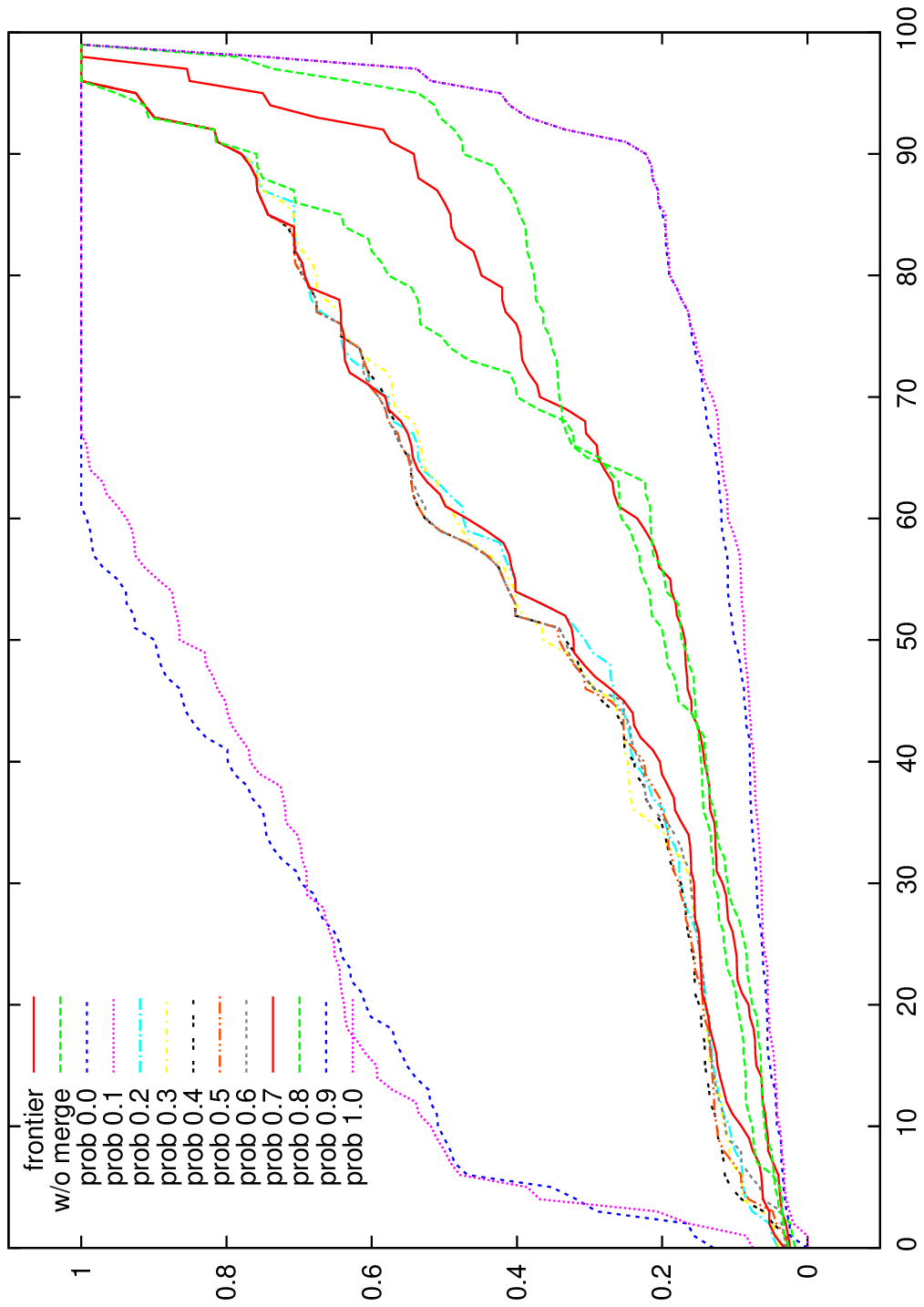}{Constrained start+goal scenario}\\
\caption{Performance comparison of two ON scenarios.
The performance of the proposed method is demonstrated for different settings $X$ of the self-localization failure rate $P^E$: ``prob $X$", as well as for two baseline methods: ``frontier" and ``w/o merge".
Vertical axis: SPL performance.
Horizontal axis: Target object ID.
Higher SPL indicates better performance.
For clarity, target object IDs are independently sorted for each individual curve.
}\label{fig:B}
\end{center}
\end{figure}
}

\newcommand{\figC}{
\begin{figure}[t]
  \begin{center}
\hspace*{5mm}\FIG{8}{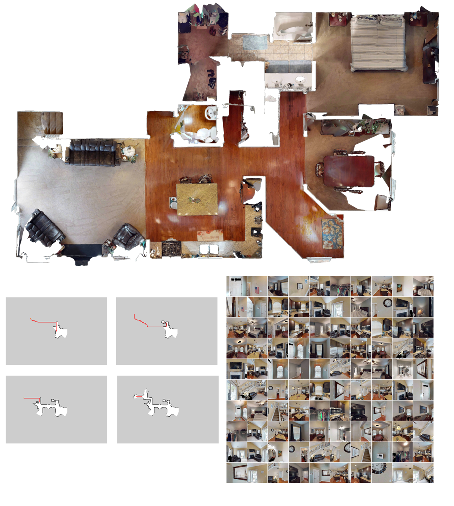}{}
\caption{
Experimental setup.
Top: A bird's-eye view of the workspace.
Bottom left: An illustration of the map generation process.
Bottom right: 100 target images.
}\label{fig:C}
  \end{center}
\end{figure}
}

\begin{abstract}
This work explores the potential of brief inter-agent knowledge transfer (KT) to enhance the robotic object goal navigation (ON) in unseen and unfamiliar environments. Drawing on the analogy of human travelers acquiring local knowledge, we propose a framework in which a traveler robot (student) communicates with local robots (teachers) to obtain ON knowledge through minimal interactions. We frame this process as a data-free continual learning (CL) challenge, aiming to transfer knowledge from a black-box model (teacher) to a new model (student).
In contrast to approaches like zero-shot ON using large language models (LLMs), which utilize inherently communication-friendly natural language for knowledge representation, the other two major ON approaches---frontier-driven methods using object feature maps and learning-based ON using neural state-action maps---present complex challenges where data-free KT remains largely uncharted.
To address this gap, we propose a lightweight, plug-and-play KT module targeting non-cooperative black-box teachers in open-world settings. Using the universal assumption that every teacher robot has vision and mobility capabilities, we define state-action history as the primary knowledge base. Our formulation leads to the development of a query-based occupancy map that dynamically represents target object locations, serving as an effective and communication-friendly knowledge representation. We validate the effectiveness of our method through experiments conducted in the Habitat environment.
\end{abstract}

\section{Introduction}

Even brief inter-human knowledge transfer (KT) can assist travelers in exploring target objects (e.g., accommodation) in unfamiliar and unseen places by acquiring local knowledge from locals. Similarly, we argue that the same applies to so-called object navigation (ON) tasks \cite{ogn1, li2022object, habitatchallenge2023}, where robots explore target objects. Specifically, in the proposed KT framework, a traveler robot (student) aims to acquire ON knowledge from local robots (teachers) encountered in unseen and unfamiliar places through brief inter-agent communication (Fig. \ref{fig:A}), thereby avoiding the labor-intensive and hazardous aspects of ON tasks themselves. We frame this as a data-free continual learning (CL) problem \cite{tsukahara2024trainingselflocalizationmodelsunseen}, which aims to transfer ON knowledge from a black-box existing model (teacher) to a new model (student).

\figA

Recently, a new approach to zero-shot ON \cite{10373065} leveraging large language models (LLM) has emerged, utilizing natural language prompts, which have excellent potential as inherently communication-friendly means of inter-agent KT.
Specifically, natural language prompts are employed as a means for robot agents to extract common-sense reasoning or zero-shot task instructions from a large inference model (LLM) agent. For example, in the LGX method \cite{10373065}, the robot first captures omnidirectional visual images, performs open-vocabulary object detection, executes task instructions based on the LLM, and achieves efficient zero-shot ON via grounding in a vision-language model. The robot composes questions in the form of concise natural language prompts that include visual, historical, or task-related information from detected objects. 
Such language prompts have great potential for inter-agent KT, although they have not yet been explored.

The other two major ON approaches,  frontier-driven ON using object feature maps \cite{Chen-RSS-23} and learning-based ON using neural state-action maps \cite{NEURIPS2020_2c75cf26}, present more complex challenges, for which data-free KT remains largely unexplored or unsolved.
For instance, in the former approach, summarizing an object map---comprising high-dimensional features like object semantics, appearance, and spatial information--- into a communication-friendly compressed format is a challenging task even for humans.
The most precise description of object maps is likely the high-dimensional feature map constructed by SLAM, but this format is not necessarily a lightweight representation suitable for transfer. Typically, the communication cost is substantial, depending on the workspace described and the datasize per unit space. Furthermore, the challenge of data-free KT of neural models is only beginning to be explored \cite{tsukahara2024trainingselflocalizationmodelsunseen}. Clearly, a more comprehensive and lightweight framework for KT is needed.

This work opens new avenues for inter-agent data-free KT in ON using the frontier-driven ON approach as a case study.
We aim to develop a minimal, plug-and-play KT module targeting black-box teachers, which may not necessarily be engaged in or cooperative with ON tasks, in general open-world environments. We introduce a minimal universal assumption---``every teacher robot possesses vision and mobility functions"---applicable to potential teacher robots, ranging from home appliances to autonomous vehicles. 
This assumption is used to define state-action history as the sole knowledge base on the teacher's side. 
We formulate the aforementioned data-free CL task as a KT problem and demonstrate that a query-based occupancy map, representing a likelihood distribution of target object locations, can be constructed on-the-fly and serves as a communication-friendly knowledge representation.
We verify the fundamental effectiveness of the proposed method through experiments using the Habitat environment \cite{habitatchallenge2023}.

\section{CON Setup}

This section extends the typical single-agent ON to the novel multi-agent ON addressed in this paper, with careful consideration to make minimal assumptions about non-cooperative, black-box, and non-ON-engaged teacher robot agents.

\subsection{ON}

The objective of the ON task is for a robot to search for a given target object, such as in the form of an image, in an unseen and unfamiliar workspace. 
This task is positioned as a variation of the classical environment exploration problem, which has been typically addressed using frontier-based exploration approaches.
There are mainly three cutting-edge approaches: 
frontier-based methods \cite{Chen-RSS-23}, 
learning-based methods \cite{NEURIPS2020_2c75cf26}, 
and 
zero-shot methods \cite{10373065}. 
Specifically, we assume a 2D indoor habitat workspace as shown in Figure \ref{fig:C}. Typical examples of target objects include furniture and small items. No prior knowledge such as the layout of the workspace is provided.

\figC

\subsection{Student}

The student robot is engaged in the ON task, following the standard ON task setup. 
This robot possesses capabilities for measurement, recognition, planning, and control for ON, which include acquiring visual images, object recognition, self-localization, odometry, local planning, global planning, collision avoidance, and path following. For details, please refer to the ON survey literature \cite{ogn1}. However, even proficient ON agents can hardly expect efficient and safe object exploration in unseen and unfamiliar workspaces when operating alone. Thus, they are compelled to rely on brute force environmental exploration or predictions based on semantic relevance.
This strongly motivates the introduction of KT from local robots (teachers) to the ON robot (student).

\subsection{Teacher}

The teacher robot is not necessarily engaged in the ON task, nor is it always cooperative with the ON task. This is what distinguishes typical multi-robot ON from our problem. Our framework actively leverages robots unrelated to the ON task, such as smart home appliances or self-driving vehicles, as teacher robots. 
A black-box teacher model is assumed, where direct access to the object feature map is not permitted, and interaction with the map is only allowed through a predefined question-answer protocol.
It is important to note that this assumption applies not only to the frontier-driven ON approach \cite{Chen-RSS-23} but also to certain frameworks within the learning-based ON approach \cite{NEURIPS2020_2c75cf26}, where the object feature map module is managed independently of planner modules such as global planners and local planners.

\subsection{Dataset}

Making too many assumptions about the qualifications of a teacher robot can lead to the exclusion of many potential teachers. Therefore, we make only an assumption that universally apply to a wide variety of potential teachers, ranging from smart appliances to self-driving vehicles:
``teachers possess datasets of view images and action histories acquired in past visual navigation." 
No special assumptions are made regarding the format of this history, for example, raw sequences of sensor data or actions can be replaced with their compressed neural representations.

\subsection{Student Proxy}

Instead of assuming KT functionality in the teacher robot, we introduce the concept of a student proxy as a plug-and-play KT module that can be installed on-demand in the teacher (Fig. \ref{fig:A}). The role of the student proxy is to access the teacher's dataset, structure the dataset, create responses to queries from the student, and send them to the student via a wireless network. 
Severely limited communication capacity is assumed for transmitting the query and response via this wireless system.

\subsection{Object Map}

The student and the (teacher-side) student proxy are capable of constructing an object map for a given target object on-the-fly from their datasets. The object map is an occupancy map, where the cell values represent the occupancy probability of the target object.

\tb{Definition: }
Our object map follows the standard formalization of occupancy grid maps but primarily differs in the probability estimation method. Occupancy grid maps have been extensively studied across various sensor modalities, such as LRF, RGBD cameras, and ultrasonic sensors, typically involving a series of steps: observation, discretization, probability estimation, and map updates. Our map diverges from traditional approaches in that, instead of using occupancy measurement values (e.g., ultrasonic time-of-flight values) to predict obstacle occupancy probability, we estimate the occupancy probability based on the similarity between the queried target object and the view images that encompass the respective cell. Semantic, visual, and spatial similarities can be used as the similarity measure.

\tb{Advantage:} 
The object map is significantly lightweight compared to the original dataset, making it communication-friendly. This can be attributed to two main reasons. First, instead of using a typical high-dimensional embedding vector (e.g., appearance, semantic, and spatial feature vectors), each cell value is a scalar similarity value or relevant score of the target object, which reduces the spatial cost per cell. Second, most cells within the grid are usually unrelated to the target object and can be ignored, and this sparseness allows for further compression of the grid map.

\tb{Implementation:} The spatial resolution of the object map is set to 0.1m. The embedding vectors use intermediate signals from VGG16. Cosine similarity is used for similarity calculation.

\subsection{Self-localization}

The student proxy has the ability to localize the viewpoint of the student's query view image in the teacher's map coordinate system, or in other words, predict the coordinate transformation from the teacher's map coordinate system to the student's map coordinate system. 
The coordinate transformation is represented by SE2, consisting of both rotation and translation, making it sufficiently lightweight.
This coordinate transformation is beneficial for merging the student and teacher object maps.

\tb{Definition:}
Self-localization is a fundamental problem in visual navigation and is a well-studied research field in both robotics and vision. Here, it is formulated as a supervised learning problem focused on training a place classifier using viewpoint-annotated view images \cite{planet}. The definition of place classes follows a typical grid partitioning approach \cite{itsc2019hiroki}, where the workspace explored by the teacher is divided into a grid, with each grid cell defined as a place class. The classification problem is formulated as an $(N+1)$-class classification problem, aiming to classify a given view image into one of the $N$ known place classes or a novel place class.

\tb{Implementation:}
The viewpoint annotations are obtained by partitioning the viewpoint trajectories derived from the teacher's history data into place classes. The supervised learning process involves fine-tuning a pre-trained VGG16 model as a place classifier. This is followed by post-processing, where, if the robot encounters an unseen viewpoint, outliers are reliably detected using post-verification methods such as RANSAC. The grid partitioning is set with a spatial resolution of 1 meter and an angular resolution of 45 degrees.

\tb{Training:}
The training process follows the standard protocol for supervised learning. Data augmentation techniques, including width and height shifts, zooming, and nearest-neighbor interpolation, are employed to enhance the diversity of images and improve the model's generalization performance. Specifically, ImageDataGenerator is utilized with a shift range of 0.2 for both width and height, and a zoom range also set to 0.2. These augmentations are applied during image loading. A global average pooling layer is included, followed by a fully connected layer with 1024 units and ReLU activation. Additionally, a 50\% dropout layer is incorporated to enhance the model's generalization capability. Finally, a softmax layer corresponding to the number of classes is added to construct the output layer. The Adam optimizer is used for compilation, with the loss function being sparse categorical cross-entropy, and model accuracy as the evaluation metric. The learning rate is set to 3e-5, with training conducted over 5 epochs and a batch size of 32.

\subsection{Map Merging}

When the outcome of self-localization is determined not to be an outlier, the teacher's map is aligned and merged with the student's map using the self-localization hypotheses as an anchor, resulting in a more informative student map.
First, a small set of $N$ self-localization hypotheses with the highest likelihood is selected ($N=5$).
Then, for each hypothesis, the teacher's map is aligned under the hypothesis, and the value of each cell in the teacher's map is merged with the corresponding cell in the student's map.
Specifically, two types of methods for merging cell values are employed: max pooling and sum pooling. The result of the former is considered the primary score, and the result of the latter is regarded as the secondary score. A multi-channel object map is generated, recording both cell values.

\section{System Integration}

This section presents a specific ON system implementation to explain how the proposed KT framework is utilized to enhance the efficiency and safety of the ON task. 
Intuitively, the teacher's experience from exploring the workspace in the past is expected to be beneficial for the student, in order to avoid labor-intensive and hazardous exploration, especially in cases where the workspace is unseen and unfamiliar to the student.

The action planning framework consists of three key modules: global action planner, local action planner, and self-localization. 
The global action planner determines the next-best-subgoal and delegates robot control to the local action planner.
The local action planner controls the robot's detailed movements until it encounters unexpected obstacles or reaches the sub-goal. The self-localization module is invoked at the start of the robot's operation and each time a sub-goal is reached. 
Each time non-outlier self-localization is achieved, the teacher's object map is aligned with the student's object map and merged, resulting in a more informative updated student map, which enhances safety and efficiency in exploration.

\subsection{Global Action Planning}
In global action planning, the value of a subgoal is the expected information gain obtained by simulating the movement and observations at each waypoint from the robot's current position to that subgoal. The information gain refers to the likelihood of detecting the target object in each cell within the observed area. The entire planning procedure is as follows.

(1) First, the viewpoint path is determined using Dijkstra's algorithm to find the shortest path from the current location to the candidate subgoal on an obstacle map. This obstacle map is a grid map formatted similarly to the object map, but its cell values reflect the probability of the presence of arbitrary objects, based on the distance to those objects measured by the moving stereo, rather than the likelihood of target objects' presence.

(2) Next, waypoints are placed at regular intervals (travel distance) along the viewpoint path. Then, by simulating the observation range at each waypoint, the set of observable cells is calculated. 

(3) Then, all observable cell sets at every waypoint are integrated along the shortest trajectory, and the observation range across the entire trajectory is computed as the union of these sets.

(4) Then, the scores of the cells within the observation range are aggregated to determine the value of each candidate sub-goal. 

(5) Finally, the candidate sub-goal with the highest score is selected. 

In the current experimental system, the value of each candidate sub-goal is determined by max pooling of the primary scores of the cells within its observation range. If multiple candidate sub-goals have the highest value, max pooling of the secondary scores is temporarily used as an additional criterion for differentiation.

\subsection{Local Action Planning}
The path to the sub-goal is planned on the obstacle map using Dijkstra's algorithm, and movement is restricted to four directions: forward, backward, left, and right. These steps are repeated until the robot reaches the target object. To ensure efficient exploration, any area within a one-meter radius of a previously visited location is considered free of the target object and will not be selected as a future sub-goal. 
If, during navigation, the robot reaches the edge of an unexplored area or the sub-goal area, the local planner is terminated, and control is returned to the global planner.

\section{Experiments}

We conducted the experiments using the ``habitat-sim" simulator (Fig. \ref{fig:C}). Habitat-sim is a high-performance photo-realistic simulator that supports agents, sensors, and a variety of 3D datasets, offering over 1,000 different scenes such as residential and commercial environments. In this study, we imported the workspace ``00800-teesavr23of" into the simulator.

\subsection{Setups}

The performance of the ON task is measured using the SPL (Success weighted by Path Length) metric. SPL assesses how efficiently the agent reaches the target, taking into account both the success rate and the efficiency of the path. It is defined as follows:
\begin{equation}
SPL = \frac{1}{n} \sum_{i=1}^{n} s_i \frac{1}{\max(p_i, l_i)},
\end{equation}
where 
$l_i$ is the length of the shortest path from the goal to the target for a given episode, 
$p_i$ is the length of the path taken by the agent during that episode, and 
$s_i$ is a binary indicator of success for that episode.

The experimental system is designed to faithfully represent the real-world robot we are developing \cite{DBLP:conf/sii/TerashimaKYL24}. However, two modifications have been made to facilitate the analysis of the experiments.
(1) To investigate the impact of self-localization error on ON performance, we replaced the self-localization model with a synthetic one, which fails with a preset constant probability: $P^E$. This model returns the classification result of an oracle self-localization model with a probability of $(1-P^E)$ and returns a random place class with a probability of $P^E$.
(2) To eliminate the effects of object detection errors and simplify the experiment analysis, we adopted an ideal object detection model. Specifically, we assumed that if the target object is within the robot's field of view, the robot will reliably detect it. The detection range was set to 1.6m.
The robot's field of view is modeled as a 3.2m radius arc with a 40-degree viewing angle.

One hundred target objects were manually selected from a workspace for use in the experiment.

\subsection{ON Senarios}

We considered the following two types of practical scenarios for the ON task.

\tb{``Constrained Start":}
The teacher robot starts from a location (with random orientation) where the target object is within its field of view. The student robot starts from a random location 3.2m away from the target object.

\tb{``Constrained Start + Goal":}
Both the start and goal locations are constrained, and specifically, the teacher's goal location is the student's start location and vice versa.

In practice, the start and goal locations that meet the above conditions may appear mid-sequence in the teacher's view sequence. In such cases, the sub-sequence before the start or after the goal can be removed to satisfy these conditions. If multiple potential teachers meet the above criteria, conducting an auction to select the best teacher would be beneficial, as we explain in Section \ref{sec:application}.

\figB

\subsection{Baselines}

To verify the effectiveness of the proposed method, we compared it with the following two methods:

\tb{``w/o merge":} 
This method uses only the student map without using the teacher map.

\tb{``frontier":}
This method randomly selects the next subgoal from the frontier regions without using object map scoring.

\subsection{Results}

Figure \ref{fig:C} shows the results of the proposed method compared to the baseline methods. The SPL exhibited high values, ranging from approximately 0.7 to 0.8. As expected, as the self-localization failure probability $P^E$  increased, the SPL gradually decreased, dropping sharply when the failure probability exceeded 20\%. Despite this, the proposed method maintained higher performance compared to other methods. 
This result demonstrates that even local robots not engaged in the ON task can serve as excellent teachers, significantly contributing to the improvement of the student's ON performance.
Even when self-localization fails, the map provided by the teacher was useful for determining appropriate subgoals, allowing the system to maintain a high SPL. However, when the failure probability exceeds 80\%, the misuse of the transferred knowledge  becomes more frequent, leading to a sharp decline in performance.

Additionally, differences in trends were observed in the two different scenarios. In the ``Constrained Start + Goal" scenario, the robot could start from a previously explored location, enabling it to use the teacher's object map from the very beginning of exploration. As a result, the performance in this scenario was higher than in the ``Constrained Start" scenario, especially when self-localization accuracy was high. However, when the success rate decreased, the merged object map of the teacher and student reflected various self-localization errors from the start, leading to a rapid decline in performance.

\section{Applications}\label{sec:application}

The findings from the experiments reveal that the proposed system is effective for the following real-world applications.

\tb{Auction:}
In this work, we addressed the continual ON-task, where a traveler robot (student) receives ON knowledge from a local robot (teacher) to improve the efficiency and safety of the ON-task in an unfamiliar and unseen workspace. Specifically, it was demonstrated that by strategically selecting a teacher equipped with object maps that include the student's view image and the target images, this KT could significantly enhance the student's ON performance. This suggests that in scenarios where multiple potential teacher robots are available, auction to actively choose a teacher robot that meets the above conditions is an effective strategy.

\tb{Asynchronization:}
In general dynamic environments, all object configurations, including the target objects, can undergo significant changes. In such environments, it is evident that the proposed approach utilizing the sensor history of the local robot can be easily applied to object tracking, monitoring, and change detection using the timestamps of the sensor data. This dynamic ON-task has been underexplored in existing research, with only a few related studies. 
In contrast to existing synchronized ON scenarios, where multiple robots collaborate to execute ON tasks simultaneously, the proposed asynchronous ON can adapt to dynamic environments by leveraging long-term historical data from the local teacher robots. In relation to this, the dual task of ON, known as active change detection, has recently been explored in \cite{DBLP:conf/sii/TerashimaKYL24}.

\tb{Verbalization:}
Inter-agent communication is expected to become increasingly important in future ON-tasks. The key to assembling natural language prompts with non-verbal knowledge as the core content lies in low-dimensionality and sparsity. Rather than comprehensively describing all objects with high-dimensional object features such as appearance, semantics, or point clouds, it is often more effective for lightweight communication to describe only the essential objects of interest to the recipient using low-dimensional descriptors such as distance or direction. In this work, it was demonstrated that even a brief message conveying the most likely location of a target object could drastically improve ON-task performance. 
Such brief messages that meet the conditions of low dimensionality and sparsity can effectively serve as core components of soft prompts.

\tb{Inversion:}
Continual learning (CL) from black-box teachers often requires solving challenging model inversion tasks. These tasks involve reconstructing pseudo-datasets that can be used to train a model with capabilities equivalent to those of the teacher. However, such inverted ON tasks have rarely been explored. 
Existing data-free KT methods have primarily focused on machine learning applications in areas like biomedical engineering, but they cannot be directly applied to SLAM or navigation tasks. 
The lightweight and efficient mapping system proposed in this work offers a plug-and-play solution for knowledge transfer (KT), applicable to black-box teachers with unknown architectures.
In line with this, a new approach to data-free KT has been explored using the fundamental problem of self-localization as a case study \cite{tsukahara2024trainingselflocalizationmodelsunseen}.

\section{Conclusions}

We explored a knowledge transfer (KT) approach for a traveling robot (student) to receive lightweight and useful knowledge from a local robot (teacher) in order to perform object navigation (ON) tasks in unfamiliar and unseen places. Unlike many existing multi-agent ON setups, we assume the existence of unfamiliar and unseen ``black-box" teachers, a universal assumption that applies to many potential robots. For non-cooperative, specification-unknown black-box teachers, we developed a plug-and-play knowledge transfer module. This module is responsible for conducting high-frequency question-and-answer sessions with the teacher on behalf of the student, extracting the necessary knowledge, compressing it compactly, and transferring it to the student. We mainly addressed two question-and-answer protocols. One focuses on localizing the target object, while the other localizes the student on the teacher's map. As a result, it was proven that lightweight and useful object maps can be transferred to the student, leading to actual improvements in the student's ON performance. We validated the effectiveness of our method through experiments conducted in the Habitat environment.

\bibliographystyle{IEEEtran} 
\bibliography{reference}

\end{document}